\documentclass[letterpaper, 10 pt, twocolumn]{ieeeconf}  

\IEEEoverridecommandlockouts                              
\usepackage[dvips]{graphicx}
\usepackage{multirow}
\usepackage{psfrag}
\usepackage{amsmath}
\usepackage{setspace}
\usepackage{color}
\usepackage{multicol}
\usepackage{lipsum}
\usepackage{tabularx}
\usepackage{rotating}
\usepackage{romannum}
\usepackage[ruled,vlined]{algorithm2e}
\usepackage{latexsym}
\usepackage{subfigure}
\usepackage{amssymb}
\usepackage{url}
\usepackage{amsmath}
\usepackage{amssymb}
\usepackage{psfrag}

\usepackage[section]{placeins}
\usepackage{arydshln}

\def\atan2{{\mbox{atan2}}}

\newcommand{\Fig}[1]{Fig.~\ref{fig:#1}}

\newcommand{\Sec}[1]{Sec.~\ref{sec:#1}}

\def\12{\frac{1}{2}}

  \def\b{\mbox{\boldmath $b$}}

 \def\j{\mbox{\boldmath $j$}}  
   \def\p{\mbox{\boldmath $p$}}
\def\q{\mbox{\boldmath $q$}}   
   \def\x{\mbox{\boldmath $x$}}
\def\y{\mbox{\boldmath $y$}} \def\z{\mbox{\boldmath $z$}}

\def\I{\mbox{\boldmath $I$}}   
   
   \def\T{\mbox{\boldmath $T$}}

\def\0{{\bf 0}}

 \def\b0{{\mbox{\boldmath $0$}}}

\def\natural{\mbox{\rm I\kern-0.2em N}}  

\def\real{\mbox{\rm I\kern-0.2em R}}  
\def\=def{\stackrel{def}{=}}

\def\beq{\begin{equation}}
\def\eeq{\end{equation}}
\def\be{\begin{equation}}
\def\ee{\end{equation}}
\def\bea{\begin{eqnarray}}
\def\eea{\end{eqnarray}}
\def\beann{\begin{eqnarray*}}
\def\eeann{\end{eqnarray*}}

\def\ds{\displaystyle}

  {\setlength{\arraycolsep}{0.14em} 
   \begin{eqnarray}}
  {\end{eqnarray}}           

\def\Beginboxit
   {\par
    \vbox\bgroup
           \hrule height1pt
           \hbox\bgroup
                  \vrule width1pt \kern2pt %
                  \vbox\bgroup\kern2pt
   }
\def\Endboxit{%
                             \kern2pt
                             \vspace*{4pt}
                       \egroup
                  \kern2pt\vrule width1pt
                \egroup
           \hrule height1pt
         \egroup
   }

\newenvironment{boxit}{\Beginboxit}{\Endboxit}
\newenvironment{boxit*}{\Beginboxit\hbox to\hsize{}}{\Endboxit}

\def\ds{\displaystyle}

 \newif\ifSmallFigure \SmallFiguretrue
 \ifSmallFigure
    \def\FIGT2{c:/usr/claudio/work/ucima/tlibro/tfigure1} 
 \else
    \def\FIGT2{c:/usr/claudio/work/ucima/tlibro/tfigure} 
 \fi

\def\sign{\mathrm{sign}}

\def\vel{\texttt{v}}

\def\vmax{\vel_{max}}

\def\acc{\texttt{a}}

\def\amax{\acc_{max}}

\def\qdm1{\q_{k-1}}
\def\vdm1{\dotq_{k-1}}
\def\acdm1{\ac_{k-1}}
\def\jdm1{\j_{k-1}}

\usepackage{hyperref}
\hypersetup{
    colorlinks=true,
    linkcolor=blue,
    filecolor=magenta,      
    urlcolor=cyan,
    pdftitle={Overleaf Example},
    pdfpagemode=FullScreen,
    }

\graphicspath{{./Figures/},{./NewFigures/},{./NewNewFigures/}}

\newcommand\red[1]{{\color{red} #1}}

\newtheorem{rem}{Remark}
\begin{document}

\title{\LARGE \bf Optimal Feed-Forward Control for\\Robotic Transportation of Solid and Liquid Materials\\via Nonprehensile Grasp }
%
%

\author{Luigi~Biagiotti,~ Davide~Chiaravalli,~Riccardo~Zanella~and~Claudio~Melchiorri %
\thanks{L. Biagiotti is with the Department of Engineering ``Enzo Ferrari'', University of Modena and Reggio Emilia, via Pietro Vivarelli 10, 41125 Modena, Italy, e-mail: luigi.biagiotti@unimore.it.}
\thanks{D. Chiaravalli,  R. Zanella and C. Melchiorri  are with the Department of Electrical, Electronic and Information Engineering ``Guglielmo Marconi'', University of Bologna, Viale del Risorgimento 2, 40136 Bologna, Italy, e-mail:  \{davide.chiaravalli\},\{riccardo.zanella2\} ,\{claudio.melchiorri\}@unibo.it.}
}
%
%

\maketitle
\thispagestyle{empty}
\pagestyle{empty}

\begin{abstract}
In everyday life, we often find that we can maintain an object's equilibrium on a tray by adjusting its orientation. Building upon this observation and extending the method we previously proposed to suppress sloshing in a moving vessel, this paper presents a feedforward control approach for transporting objects with a robot that are not firmly grasped but simply placed on a tray.
The proposed approach combines smoothing actions and end-effector re-orientation to prevent object sliding. It can be integrated into existing robotic systems as a plug-in element between the reference trajectory generator and the robot control. To demonstrate the effectiveness of the proposed methods, particularly when dealing with unknown reference signals, we embed them in a direct teleoperation scheme. In this scheme, the user commands the robot carrying the tray by simply moving their hand in free space, with the hand's 3D position detected by a motion capture system.
Furthermore, in the case of point-to-point motions, the same feedforward control, when fed with step inputs representing the desired goal position, dynamically generates the minimum-time reference trajectory that complies with velocity and acceleration constraints, thus avoiding sloshing and slipping.
More information and accompanying videos can be found at \url{https://sites.google.com/view/robotwaiter/}.
\end{abstract}

\section{Introduction}
\begin{figure}[t!]
	\centering
	\includegraphics[width=\columnwidth]{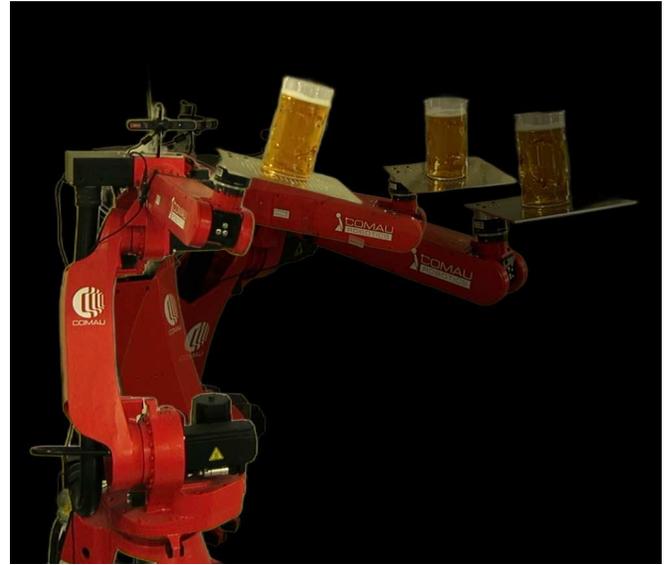}
	\caption{\label{fig:ComauBeer} Industrial robot manipulating a mug of beer resting on a tray (picture from the video \cite{Moriello2018VideoIROS}).}
\end{figure}

As highlighted in a recent survey on Nonprehensile Dynamic Manipulation \cite{Ruggiero18}, a typical example of a non-prehensile task performed by humans consists of "carrying a glass full of liquid on a tray." This is exactly the task that we aim to replicate in this research activity by means of a robotic manipulator as shown in \Fig{ComauBeer}. In fact, the relocation of an object from position A to position B can be safely performed by grasping the object and moving it, but it requires a mechanism, such as a gripper, capable of restraining the object. In many cases, dynamic nonprehensile manipulation may offer advantages \cite{Lynch99,Ruggiero18}, as there is no need to firmly grasp the object, eliminating the requirement for grasping mechanisms and reducing the risk of damaging the object with excessive grasping forces. However, several limitations arise, such as bounds on the maximum velocities and accelerations that can be applied to the robot to maintain the object on the carrying structure.

The problem becomes even more complicated when the object to be manipulated without a firm grasp is represented by a liquid contained in a vessel. In this case, the challenge of balancing the object is combined with the need to suppress the sloshing of the liquid, which could cause it to spill. Although this type of problem is considered an example of non-prehensile manipulation, to the best of our knowledge, a robot-based solution has not yet been proposed in the literature. In fact, the existing literature focused on sloshing suppression typically involves firmly connecting the vessel to the robot flange or the manipulation mechanism.

This paper builds upon the results reported in the conference paper \cite{Biagiotti18}, which are briefly summarized in the section describing the experimental results, and generalizes the proposed control method to eliminate the need for a stable connection between the container and the robot flange. As a cascaded result also the problem of minimum-time transportation of solid objects via nonprehensile grasp is solved. 
\section{Related works and  contribution of the paper }
This paper joins two research lines, such as sloshing suppression and nonprehensile manipulation, therefore it is necessary to take into account both research fields. We initially consider them separately but in the paper, a tight connection between the two problems will be proved from an analytical and an experimental viewpoint.\\
As mentioned above, this paper extends the approach proposed in \cite{Biagiotti18}, whose basic idea consists of combining  a smoothing action applied to the desired trajectory for suppressing the oscillations of the liquid, and a compensation of the lateral accelerations based on a proper modification of the container’s orientation. 
 The proposed approach is a typical example of feed-forward control providing the proper reference trajectory for the robot/machine on the basis of the model of the slosh dynamics. With respect to feedback methods that rely on a measure of the liquid surface configuration, feed-forward algorithms offer  some advantages since they do not require an additional sensor apparatus (that in case of detection of the liquid configuration, especially in containers that are not specifically designed for the robotic applications like e.g. a glass or a bottle,  may be quite complicated)
 and can be easily implemented in standard industrial robots/machines, without modifications of their control systems. For this reason, the techniques   based on feed-forward are the most common solutions in the field of liquid manipulation. For instance, in  \cite{Feddema1997} and \cite{Chen2007}  the of slosh dynamics is compensated via tilt angle modification. Input shaping techniques are widely used in conjunction with smooth trajectory planning and other kinds of filtering/smoothing methods, see  \cite{terashima1996,ABOELHASSAN20091,Pridgen2013,aribowo2015,Zang2015} among many others.  Alternative feed-forward methods are based on the optimization of the reference trajectories applied to the liquid container, computed by taking into account the dynamic model of the system and a number of constraints, like maximum velocity, acceleration, or even the location of possible obstacles \cite{yano2001,Yano2005,consolini2013,Maderna2018,Reinhold19,Muchacho22}.\\
 The main drawback of all these methods is obviously the poor robustness of the system with respect to unmodelled dynamics (in this specific case higher order sloshing modes) and inaccurate knowledge of the physical parameters. For this reason,  some authors combine the feed-forward
 control with a feedback compensation, e.g. based on $H_\infty$ loop shaping methods \cite{terashima1996,TERASHIMA2001607}, or
 mix different feed-forward algorithms like input shaping and smoothing filters \cite{Zang2015}, or input shaping and tilting compensation, like in our previous paper \cite{Moriello2017}. \\
By analyzing the literature on the transportation of solid objects using robotic systems without grasping mechanisms, it becomes evident that the methods employed are highly similar to those utilized for suppressing sloshing. Firstly, the majority of procedures aimed at preventing object slippage on a tray manipulated by a robot rely on feed-forward approaches. The simplest methods involve effectively planning the timing and motion profile along the desired path to minimize the overall trajectory duration while ensuring that the contact forces remain within the friction cones \cite{Lertkultanon2014,Luo2017,Csorvasi2017}.  
However, when implementing these methods on real robots, unmodeled effects such as incorrect estimation of the friction coefficient can lead to failures or, conversely, to over-conservative trajectories. To address this, other works have proposed strategies to more robustly avoid object slippage by adjusting the tray orientation and defining 6D trajectories.
In \cite{Martucci2020}, the orientation of the end-effector is adjusted to ensure that the inertial forces acting on the grasped object always remain within the Grasp Spatial Force Space (GSFS), which represents the range of forces that a grasp can withstand. It should be noted that defining the GSFS relies on the contact model between the object and the end-effector, as well as accurate knowledge of frictional parameters. Additionally, the proposed approach is based on an offline optimization method and cannot be utilized in applications where the reference position is not known in advance.
Similar considerations are applicable to the approach presented in \cite{GATTRINGER2022}, where a minimum-time trajectory is sought using a dynamic programming algorithm applied to a chain of integrators, subject to constraints on internal variables (e.g., velocity and acceleration) and tangential force affecting the object. Notably, the procedure in \cite{GATTRINGER2022} bears striking resemblance to the one employed in \cite{Reinhold19} for sloshing suppression.
A comparable concept is explored in \cite{Selvaggio2022}, where the orientation of the tray supporting the object is dynamically adjusted to increase the distance between the contact forces and the boundaries of the friction cone. However, in this case, the procedure operates in real-time and is implemented within a direct tele-manipulation architecture with shared control to maintain object stability.\\
In \cite{Subburaman2023}, an algorithm similar to the one that we proposed for sloshing suppression in \cite{Biagiotti18}  is utilized when high accelerations are induced by translational motion. Specifically, the horizontal motion is decomposed into the $x$ and $y$ directions, and a rotation $\theta_x$ around the $x$-axis is imposed on the tray, which depends on the acceleration $a_y$ along the $y$-axis (and vice-versa):
\be
\theta_x = \tan^{-1}\left( \frac{\mu g - a_y}{g + \mu a_y}\right)
\label{eq:ThetaCompensationMu}
\ee
where $\mu$ represents the static friction coefficient between the object and the tray. Two main differences distinguish the angle in (\ref{eq:ThetaCompensationMu}) from the tilting compensation proposed in \cite{Biagiotti18}. First, it appears that vertical acceleration does not influence the angle $\theta_x$. Second, the friction coefficient $\mu$ plays a role in its computation. As a side effect, when $a_y=0$ a residual compensation angle $\theta_x = \tan^{-1}(\mu)$ remains.\\
The first question that arises when analyzing all the aforementioned approaches for non-prehensile transportation, especially the last one, is whether an expression of feed-forward compensation can be found that does not depend on friction parameters. This doubt arises because the approach we proposed for sloshing suppression, based on tilting compensation, only requires knowledge of linear accelerations \cite{Biagiotti18}.An experimental application to a waiter task suggests that the same control is able to maintain an object in equilibrium on a tray even when very fast motions are applied (see the video \cite{Moriello2018VideoIROS}). Additionally, while many researchers agree on the application of orientation compensation to prevent slipping, the problem of determining the optimal location of the center of rotation has not been addressed yet. \\
An answer to the two above-mentioned problems is the main contribution of the first part of this work, by considering both liquid and solid objects/materials. In \Sec{model},  a detailed 2D analytical model is used to derive the optimal feed-forward angular compensation in the general case of a liquid contained in a vessel lying on a flat tray, which is moved by a robot manipulator. The compensation of the motion of the tray does not require any knowledge about the system model (and, in particular, the friction coefficient), but only the estimation of the imposed accelerations. For this reason, in \Sec{Smoothers}, a mechanism based on smoothers is proposed for generating minimum-time point-to-point motions or filtering external reference trajectories while deriving the related accelerations without explicit differentiation. The overall architecture of the feed-forward system for nonprehensile 3D manipulation with sloshing suppression capabilities is presented in \Sec{feedforward}, and proper choices of free parameters (order and coefficients of the smoothers, location of the center of rotation, etc.) are recommended according to different application scenarios, involving solid/liquid materials and point-to-point/multi-point trajectories. Finally, in \Sec{model}, the proposed approach is experimentally validated in all the aforementioned scenarios. In particular, to demonstrate its effectiveness when dealing with unknown trajectories, a simple teleoperation task has been set up. In this task, the user directly commands the robot by moving its hand in free space, and the 3D position of the hand is detected by a motion capture system. The robot, carrying a pot filled with liquid, then tracks the motion generated by the devised algorithm.

\section{A model-based approach for the synthesis of  slosh-free and sliding-free motions}\label{sec:model}
\begin{figure}[t!]
	\centering
	\psfrag{D}[c][c][0.9][0]{ $d_z$}
	\psfrag{h}[c][c][0.9][0]{ $h$}
	\psfrag{l}[c][c][0.9][0]{ $l$}
	\psfrag{m}[c][c][0.9][0]{ $m$}
	\psfrag{M}[c][c][0.9][0]{ $M$}
	\psfrag{b}[c][c][0.7][0]{ $\beta$}
	\psfrag{t}[c][c][0.8][0]{ $\theta$}
	\psfrag{x}[c][c][0.9][0]{ $x$}
	\psfrag{z}[c][c][0.9][0]{ $z$}
	\psfrag{X}[c][c][0.8][0]{ $x_c$}
	\psfrag{Z}[c][c][0.8][0]{ $z_c$}
	\psfrag{W}[c][c][0.8][0]{ $x_t$}
	\psfrag{T}[c][c][0.8][0]{ $z_t$}
	\psfrag{d}[c][c][0.8][0]{ $d_x$}
	\includegraphics[width=\columnwidth]{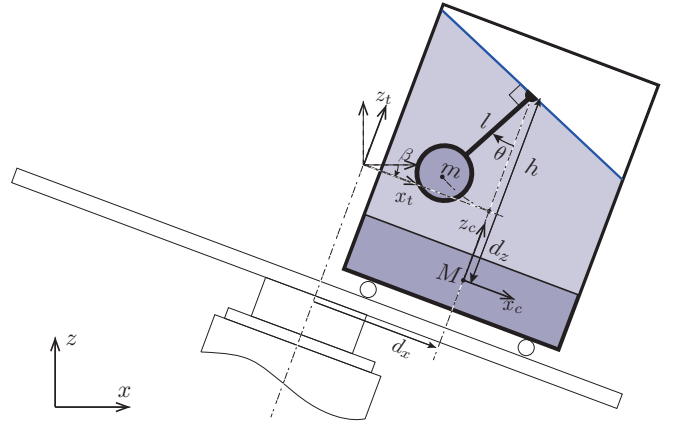}
	\caption{\label{fig:singlepend_mod} Simplified mechanical model approximating the first asymmetric sloshing mode in a cylindrical vessel.}
\end{figure}

To explain the proposed control approach, let's consider a simplified scenario in which a container, possibly filled with liquid, rests on a flat tray that is connected to the robot flange. The flange can translate on the $x-z$ plane and rotate around an axis that is perpendicular to this plane. As a result, the behavior of the system can be described with a planar model. The goal is to keep the container in its initial location and minimize sloshing phenomena while the tray is moved in an arbitrary way. Therefore,  the Cartesian position $(x_t,y_t)$ of the tray and the related kinematic quantities such as velocity and acceleration are the disturbance inputs acting on the system, while the tilting angle $\beta$ is the only manipulable input, as shown in \Fig{singlepend_mod}.

The dynamics of the liquid in the moving vessel is often modeled with an equivalent mechanical model consisting of a rigid mass $m_0$ and a series of pendulums with mass $m_j$, length $l_j$, and support points located at a distance $L_j$ from the undisturbed free surface of the liquid \cite{ibrahim2005liquid}.
For control purposes, the model is further simplified by considering only the first asymmetric mode of the slosh, that is a single pendulum with mass $m$, length $l$, and pivot located at the center of the liquid surface. Moreover, it is assumed that the pendulum is always orthogonal to the liquid surface which is supposed to be flat. Therefore, the model of the sloshing dynamics inside a container moving on the plane can be represented as in \Fig{singlepend_mod}. 

Additional simplifying  assumptions involve the container and the mass of liquid which is not affected by sloshing phenomena.
These masses are lumped in the center of the point mass that in \Fig{singlepend_mod} is denoted by  $M$. It is supposed that this mass never loses contact with the plate (this hypothesis is equivalent to the assumption that the normal forces exerted by the container on the plate are always positive) and cannot roll on its surface.

The position of the (reference frame attached to the) container is
\begin{align}
x_c = &x_t + d_x \cos(\beta)- d_z \sin(\beta)\nonumber\\
z_c = &z_t + d_x \, \sin(\beta) +d_z \cos(\beta) \nonumber
\end{align}
while the position of the mass of the pendulum is given by
\begin{align}
x_m = &x_t + d_x \cos(\beta) - h \sin(\beta)+l\sin(\beta+\theta)\nonumber\\
z_m = &z_t + d_x\, \sin(\beta) + h \cos(\beta)-l\cos(\beta+\theta)\nonumber
\end{align}
where the pair $(x_t,y_t)$ denotes the position of the tray and $\beta$ the tilting angle with respect to the base reference frame, $(d_x,d_z)$ is the position of the container with respect to the tray,  $h$ is the height of the liquid's surface in the frame of the tray,  and finally $l$ and $\theta$ denote the length and the rotation angle of the pendulum, respectively.\\
The equations describing the dynamics of the vessel and of the pendulum can be derived by using the Lagrange equations. The kinetic energy of the overall system composed of the container with the still liquid ($M$) and the oscillating mass $m$ representing the sloshing, is given by

{\small
\begin{align}
T =& \frac{1}{2}M \big(\dot x_{c}^2+\dot z_{c}^2 )  + \frac{1}{2}m \big(\dot x_{m}^2+\dot z_{m}^2  \big)\nonumber\\ = &
\frac{1}{2} M \left(\left( -(d_x \sin (\beta)+ d_z \cos (\beta))\dot \beta+\cos (\beta) \dot d_x +\dot x_t\right)^2\right.\nonumber\\&\left.+\left((d_x \cos (\beta)-d_z \sin (\beta))\dot \beta+\sin (\beta) \dot d_x +\dot z_t\right)^2\right)\nonumber \\& + \frac{1}{2}m \left(\left(-\left(d_x  \sin (\beta)+h \cos (\beta)\right)\dot \beta + l \left(\dot \beta +\dot \theta\right) \cos (\beta+\theta)\right.\right.\nonumber\\&\left.+\cos (\beta) \dot d_x +\dot x_t\right)^2 +\left(\left( d_x  \cos (\beta)-h  \sin (\beta)\right)\dot \beta \right.   \nonumber \\ &\left.\left.+l (\dot \beta +\dot \theta ) \sin (\beta + \theta)+\sin (\beta) \dot d_x +\dot z_t\right)^2\right)\nonumber
\end{align}
}
while the potential energy is equal to
\begin{align}
V =&  M\,g\,z_c + m\,g\,z_m \nonumber\\ \nonumber
  =&  M\,g\,(z_t+d_x\sin(\beta) +d_z\cos(\beta))  \\&+m\,g\, (z_t+ d_x\sin(\beta)+h\cos(\beta)-l\cos(\beta+\theta))\nonumber
\end{align}
where $g$ is the gravity acceleration. The Lagrangian can be computed as $\mathcal{L}=T-V$.

\subsection{Sloshing dynamics model and compensation}
\label{ssec:SloshingModel}
By considering the Lagrange equation with respect to $\theta$, the differential equation describing the dynamics of the pendulum is obtained, i.e.
\begin{align}
 l\, \ddot \theta +\frac{b_{lc}}{m l}\dot \theta + \Big(l  - h \cos (\theta) + d_x \sin(\theta) \Big) \ddot \beta\nonumber \\ +\cos(\theta)\left(-d_x \dot \beta^2+\ddot d_x\right)+\sin (\theta) \left(2 \dot \beta \dot d_x-h \dot \beta^2\right)\nonumber \\+\sin (\beta+\theta) \left(g+\ddot z_t \right)+ \cos (\beta+\theta)\,\ddot x_t&= 0 \label{eq:PendDynamics}
\end{align}
where the nonconservative term $\frac{b_{lc}}{m l} \dot \theta$ takes into account the damping force between the liquid and the container.
Note that  (\ref{eq:PendDynamics}) is exactly the same equation found in \cite{Feddema1997}, but the control solution that will be deduced to maintain $\theta$ to zero is rather different from the one proposed here.
By assuming that the position of the container on the tray  does not change during the manipulation task (and is initially null), i.e. $d_x = \dot d_x = \ddot d_x = 0$, equation (\ref{eq:PendDynamics}) becomes
\begin{align}
 l\, \ddot \theta +\frac{b_{lc}}{m l}\dot \theta + \overbrace{\left(l  - h \cos (\theta)\right) \ddot \beta\nonumber -h\sin (\theta)\dot \beta^2}^{u_r}\nonumber \\+\underbrace{\sin (\beta+\theta) \left(g+\ddot z_t \right)+ \cos (\beta+\theta)\,\ddot x_t}_{u_t}&= 0. \label{eq:PendDynamics1}
\end{align}
where $u_r$ and $u_t$ are the external disturbances caused respectively by the rotation and the translation motions imposed to the container.
Consequently, in order to prevent  oscillations, that model the sloshing of the liquid surface, it is necessary to enforce   $(\theta(t), \dot \theta(t))^T=(0,0)^T$, $\forall t\ge t_0$ to be an asymptotically stable equilibrium state  for the second order system (\ref{eq:PendDynamics1}). Since the accelerations $\ddot x_t$ and $\ddot z_t$ are set by the application, being the consequence of the translational trajectory imposed to the tray during the manipulation task, this can be only achieved by acting on the rotation angle $\beta$ of the tray (consequently, $\dot \beta \neq0$ and $\ddot \beta \neq0$). In particular, to nullify $u_t$ and $u_r$ it is necessary to assume:
\begin{align}
u_t = 0\,\,\,\,\, (\theta=0)&\,\,\,\,\,  \Rightarrow&\beta^\star &= -\tan^{-1}  \left(\frac{\ddot x_t}{g+\ddot z_t}\right)\label{eq:AngleOfRotation}\\
u_r = 0\,\,\,\,\, (\theta=0)&\,\,\,\,\,  \Rightarrow&l &=h.\label{eq:CenterOfRotation}
\end{align}
\begin{rem}
\label{rem:PendAlign}
Conditions (\ref{eq:AngleOfRotation}) and (\ref{eq:CenterOfRotation})  have a straightforward physical interpretation. The condition (\ref{eq:AngleOfRotation}) describes the compensation of the lateral  accelerations applied to the container by imposing a rotation around the point located in the center of pendulum mass by an angle that aligns the container itself with the pendulum configuration that would be reached without tilting compensation\footnote{When $\ddot \beta = \dot \beta = \beta = 0$, the equilibrium point of (\ref{eq:PendDynamics1}) becomes \[(\theta, \dot \theta)^T=\left( -\tan{-1}  \left(\frac{\ddot x_t}{g+\ddot z_t}\right),0\right)^T.\]}, while (\ref{eq:CenterOfRotation}) aims at suppressing the effects of the consequent angular acceleration $\ddot \beta$ imposed to the container.
\end{rem}

It is worth noticing that the control is based on a feed-forward compensation of external disturbances that cause sloshing. However,  this compensation only depends on the estimations of $\ddot x_t$ and $\ddot z_t$ (while specific parameters of the container/liquid do not appear in (\ref{eq:AngleOfRotation})) and is, therefore, free from typical issues of feed-forward control related to modelling errors.\\

If the container cannot be rotated, i.e. $\ddot \beta = \dot \beta = \beta = 0$, and doesn't move with respect to the tray, the equation describing the sloshing dynamics becomes
\be
l\ddot \theta+ \frac{b}{m l} \dot \theta + \sin(\theta)\big( g+\ddot z_t\big) +\cos(\theta)\ddot x_t = 0.
\label{eq:PendDynamicsNoBeta}
\ee
By linearizing (\ref{eq:PendDynamicsNoBeta}) for $\theta = \dot \theta = 0$ and $\ddot x_t = \ddot z_t = 0$ the  model
\be
l \ddot\theta+ \frac{b}{m l} \dot\theta  + g \theta= - \ddot x_t
\label{eq:PendDynamicsLinear}
\ee
can be deduced. It is a second-order system
\be
\ddot \theta  +2\delta\omega_{n}\,\dot \theta+ \omega_{n}^2\theta = u
\label{eq:SloshingLinMod}
\ee
whose parameters $\omega_{n}$ and $\delta$ depend on the characteristics of liquid and container, and the input $u$ is proportional to the  accelerations along the $x$ axis. 
In this case, it is not possible to compensate for the external acceleration $\forall t$, but  according to a typical approach for residual vibration suppression in mechanical systems, it is possible to reduce the  pendulum swing at the end of motion by shaping the input $u$ with a proper filter \cite{Pridgen2013}. To this purpose, it is necessary to know, via analytic or estimation methods, the characteristic parameters $\omega_n$ and $\delta$ of the sloshing phenomenon.

\subsection{Container's dynamic model and sliding compensation}
The Lagrange equation with respect to $d_x$ provides the equation describing the sliding dynamics of the container, full of liquid, on the tray, i.e.
\begin{align}
(m+M) \ddot d_x+ b_{ct} \dot d_x+(l \cos (\theta)-h)m\ddot \beta + m\,l \cos (\theta)  \ddot \theta  \nonumber\\ - d_z M  \ddot \beta -l\, m \sin (\theta) \left(\dot \beta + \dot \theta\right)^2-(m +M) d_x \dot \beta^2  \nonumber \\ + (m+M) \underbrace{\Big(\sin (\beta) \left(g+\ddot z_t\right)+\cos (\beta) \ddot x_t \Big)}_{\,\,\,\,\,\,\,\,\,\,\,\,\,\,\,\,\,u_t \,\,(\theta=0)}&=0
\label{eq:ContainerDynamics}
\end{align}
where the nonconservative term $b_{ct} \dot d_x$ takes into account the friction between the container and the tray. Note that the main source of external disturbance which triggers the motion of the container is given by the same signal $u_t$ affecting the sloshing dynamics (with $\theta=0$) multiplied by the total mass $m+M$. Therefore, condition (\ref{eq:AngleOfRotation}) assures that this contribution always equals zero.\\
Equations (\ref{eq:PendDynamics}) and (\ref{eq:ContainerDynamics}) describe the dynamics of the overall system composed of container and liquid. Despite the assumptions (\ref{eq:AngleOfRotation}) and (\ref{eq:CenterOfRotation}),   it is straightforward to verify that  $\x^T =  (d_x, \dot d_x,  \theta, \dot \theta )^T=(0,0,0,0)^T$ is not an equilibrium state for the whole fourth-order system. This is due to the fact that, if\footnote{Note that $l=h$ and $d_z =0$ are not consistent in any case.} $d_z\neq0$, the term $- d_z M  \ddot \beta$ in (\ref{eq:ContainerDynamics}) cannot be compensated in any way when a rotation of an angle $\beta$ is applied, and accordingly $\ddot \beta\ne0$. However, practical experience suggests that a container located on a (moving) flat surface can remain fixed in the initial location for moderate angular velocities/accelerations. The reason for this mismatch between practice and theory is due to the lack in the lagrangian equation (\ref{eq:ContainerDynamics}) of a term taking into account dry friction. For this reason, it is convenient to consider the dynamics of the container on the tray as a differential inclusion \cite{Wouw2004,BIEMOND2012}, i.e.
\begin{align}
(m+M) \left(\sin(\beta) \left(g+\ddot z_t\right)+\cos (\beta) \ddot x_t+\ddot d_x\right) + b_{ct} \dot d_x \nonumber \\+(l \cos (\theta)-h)m\ddot \beta + m\,l \cos (\theta)  \ddot \theta  - d_z M  \ddot \beta\nonumber \\ -l\, m \sin (\theta) \left(\dot \beta + \dot \theta\right)^2-(m +M) d_x \dot \beta^2 \in -F_s\,\sign(\dot d_x)
\label{eq:ContainerDynamics1}
\end{align}
where
\be \sign (x) = \left\{\begin{array}{ll} -1&x<0,\\
 {[-1\,\, 1]}&x=0,\\
1&x>0,\end{array} \right.
\ee
and $F_s$ is the maximum magnitude of the static friction, that depends on the friction coefficient $\mu$ and on the normal inward force to the tray surface, i.e.
\begin{align}
F_s =& \mu   \,\Big[ (M+m) \Big(\cos(\beta) \left(g+\ddot z_t\right)- \sin(\beta)\ddot x_t \nonumber\\& \left. \left.+ d_x \ddot \beta + 2\dot \beta \dot d_x -d_z\dot \beta^2 \right)+ m\left( l\, \sin (\theta) \left(\ddot \beta + \ddot \theta\right)\right.\right. \nonumber \\ &\left. \left.+l\, \cos (\theta) \dot \theta\left(2\dot \beta + \dot \theta\right)+\dot \beta^2(l \cos (\theta)-h)\right) \right].
\end{align}
The dynamic system described by (\ref{eq:PendDynamics}) and (\ref{eq:ContainerDynamics1}), with the conditions (\ref{eq:AngleOfRotation}) and (\ref{eq:CenterOfRotation}), has a unique equilibrium state given by $\x=\0$  as long as
\[\big|d_z M  \ddot \beta \big|\le F_s.\]
Unfortunately, without feedback control, the sliding dynamics of the container is not asymptotically stable. Therefore, it is necessary not to exit from $d_x = \dot d_x = 0$, even if occasionally $\theta\neq 0$ and $\dot \theta\neq 0$. This is possible if
\begin{align}
\Big|(m+M) \Big(\sin(\beta) \left(g+\ddot z_t\right)+\cos (\beta) \ddot x_t\Big)  +(l \cos (\theta)-h)m\ddot \beta \nonumber \\ + m\,l \cos (\theta)  \ddot \theta- d_z M  \ddot \beta -l\, m \sin (\theta) \left(\dot \beta + \dot \theta\right)^2 \Big| \le F_s. \label{eq:FrictionConstraint}
\end{align}
The previous condition clarifies the importance of the assumptions (\ref{eq:AngleOfRotation}) and (\ref{eq:CenterOfRotation}) for maintaining the container in its initial position. In particular, since $M\gg m$, the knowledge or the correct estimation of the translational accelerations becomes fundamental for imposing (\ref{eq:AngleOfRotation}) and consequently (\ref{eq:FrictionConstraint}). Note that the choice of rotation $\beta^\star$ in (\ref{eq:AngleOfRotation}) not only minimizes the left-hand side of the equation (\ref{eq:FrictionConstraint}) but also maximizes the value of $F_s$. As a matter of fact, the derivative of the first term in the expression of $F_s$, i.e.
\begin{align}
\frac{d}{d\beta}  \left(\mu   \, (M+m) \big(\cos(\beta) \left(g+\ddot z_t\right)- \sin(\beta)\ddot x_t \big)\right)=  \nonumber \\= \mu   \, (M+m)\,u_t \,\,\,\,(\mbox{with $\theta=0$}),\nonumber
\end{align}
is null for $\beta=\beta^\star$. Therefore, the orientation angle $\beta^\star$ is the best solution for compensating the effect of lateral acceleration also on the container and not only on the liquid dynamics.

\subsection{Model of a solid object on the tray and sliding compensation}
A particular case, though very relevant for applications, arises when only the container is considered. The resulting  dynamics, that models any solid object of mass $M$ transported by a robotic system without any grasping mechanism, can be deduced from (\ref{eq:ContainerDynamics1}) by assuming $m=0$, i.e.
\begin{align}
M \ddot d_x+ b_{ct} \dot d_x  - d_z M  \ddot \beta -M d_x \dot \beta^2  \nonumber \\ + M \Big(\sin (\beta) \left(g+\ddot z_t\right)+\cos (\beta) \ddot x_t \Big)&\in -F_s\,\sign(\dot d_x)
\label{eq:ObjectDynamics}
\end{align}
with
{\small
\[ F_s = \mu   M \!\left(\cos(\beta) \left(g+\ddot z_t\right)- \sin(\beta)\ddot x_t + d_x \ddot \beta + 2\dot \beta \dot d_x -d_z\dot \beta^2 \right).
\]
}
In this case, by imposing (\ref{eq:AngleOfRotation}) the equilibrium point becomes an equilibrium set defined by
\be
\xi = \{(d_x, \dot d_x)^T = (\bar d_x, 0)^T,  |d_z M  \ddot \beta +M \bar d_x \dot \beta^2| \le F_s\}.
\label{eq:EquilibriumSet}
\ee
Even if this set is not attractive, the system's state remains in this set  as long as the angular motion is characterized by bounded velocity $\dot \beta$ and acceleration $\ddot \beta$, so they are compliant with the constraint in (\ref{eq:EquilibriumSet}). Because of the computation of the optimal tilting angle $\beta^\star$ based on (\ref{eq:AngleOfRotation}), the angular acceleration will be bounded only if $\ddot x_t(t), \ddot z_t(t) \in \mathcal{C}^1$ and accordingly $x_t(t), z_t(t) \in \mathcal{C}^3$.  Based on the above considerations, it is possible to deduce the following conclusions.
\begin{rem}
\label{rem:JerkContinuity}
The compensation of lateral acceleration  by means of the tilting angle (\ref{eq:AngleOfRotation}) is feasible only if the translational motion imposed on the tray has a continuous jerk.
\end{rem}
\begin{rem}
\label{rem:CoR}
 To assure the compliance with the inequality condition in (\ref{eq:EquilibriumSet}), it is convenient to impose $\bar d_x=d_z=0$, i.e. locate the center of rotation in the center of mass of the object, so that this condition is satisfied $\forall \mu$ and $\forall \dot \beta,\, \ddot \beta <\infty$.\\
\end{rem}

In the absence of a tilting control, the model of a solid object sliding on the tray becomes
\begin{equation}
M \ddot d_x+ b_{ct} \dot d_x +M \ddot x_t \in -F_s\,\sign(\dot d_x)
\end{equation}
with
\[ F_s = \mu   M \left(g+\ddot z_t\right).
\]
Accordingly, the equilibrium set is
\[
\xi = \{(d_x, \dot d_x)^T = (\bar d_x, 0)^T,  |\ddot x_t| \le \mu    \left(g+\ddot z_t\right),\,g+\ddot z_t >0\}.
\]
\begin{rem}
\label{rem:FrictionLimit}  To maintain the object in its initial position without tilting compensation it is necessary to limit the lateral acceleration of the tray so that the ratio $|\ddot x_t| /   \left(g+\ddot z_t\right)$ does not exceed the friction coefficient $\mu$.
\end{rem}

\section{Smoothers and reference trajectory generation/filtering}
\label{sec:Smoothers}

As illustrated in Sec. \ref{sec:model}, the transportation of both solid and liquid materials requires minimizing the lateral accelerations imposed on the tray to avoid sloshing of the liquid and sliding of solid objects. Additionally, it may be useful to shape the spectrum of the reference trajectory to cancel possible residual oscillations of the equivalent pendulum that models the liquid dynamics. Finally, to implement tilting compensation, it is necessary to know the instantaneous values of the Cartesian accelerations imposed on the tray and guarantee that these values are bounded. All these goals can be achieved by using the so-called smoothers for trajectory generation/filtering.

A smoother is a filter with an impulse response of finite duration $T$, i.e.
\bea
h(t) &\!=\!&\left\{ \begin{array}{ll}
\ds \eta(t) \hspace{10mm} & \hspace{10mm} \mbox{if } 0 \le t \le T \\[4mm]
 0 & \hspace{10mm} \mbox{otherwise}\end{array}\right.\label{eq:ImpulseSmoother}
\eea
where $\eta(t)$ is a function, that in the simplest case assumes a constant value
\be
\,\,\,\,\,\,\,\,\,\eta(t)\,\,\, = \,\,\,\frac{1}{T} \hspace{20mm}(\mbox{\it rectangular smoother})\label{eq:RectangularFunction}
\ee
while in other, more complex, cases is based on trigonometric functions, i.e.
\be
\eta(t)\,\,\, = \,\,\,\frac{\pi}{2T}\sin \left(\frac{\pi}{T}t \right) \hspace{6mm}(\mbox{\it harmonic smoother})\label{eq:HarmonicFunction}
\ee

By Laplace transforming (\ref{eq:ImpulseSmoother}) with (\ref{eq:RectangularFunction}) the transfer function of the {\it rectangular smoother},
\be H(s)= \frac{1}{T}\frac{1-e^{-sT}}{s}, \label{eq:RectangularSmoother}\ee
and of the {\it harmonic smoother},
\be
H(s) =  \frac{1}{2}\left(\frac{\pi}{T}\right)^2\frac{1+e^{-sT}}{s^2+\left(\ds \frac{\pi}{T}\right)^2}\label{eq:HarmonicSmoother}
\ee
are obtained.
Note that for all types of smoothers, it is required that:

\[\int_0^T \eta(t)dt = 1\]

This is done to ensure that the DC gain of the corresponding filter is equal to one. Furthermore, basic considerations regarding the convolution product between the impulse responses defined in (\ref{eq:ImpulseSmoother}) and the input signal, which is assumed to be of class $\mathcal{C}^{n}$, suggest that the filtered signal will be of class $\mathcal{C}^{n+1}$ when a rectangular smoother is applied and $\mathcal{C}^{n+2}$ when a harmonic smoother is used.

The order of the smoother, which coincides with the degree of the polynomial in $s$ at the denominator, describes the capability of increasing the smoothness level of the filtered signal. Therefore, the rectangular smoother is a first-order filter and the harmonic smoother is a second-order filter. Additionally, these basic elements can be combined in a cascade configuration to obtain higher-order filters. Interestingly enough, the composition of two rectangular smoothers leads to the so-called ``trapezoidal smoother,'' which is another type of second-order smoother characterized by a trapezoidal impulse response.
Second-order smoothers are the basic tools used in this work since they ensure that the reference trajectory has limited acceleration even in the case of discontinuous input signals, such as step functions. Moreover, for a given input, they provide the first two derivatives along with the filtered output.\\
Note that when the smoother is fed by a step input, the impulse response coincides with the velocity profile of the output signal. Accordingly, the harmonic smoother yields a standard harmonic motion, while the trapezoidal smoother produces a  trapezoidal velocity trajectory.
By setting the value of $T$ of the harmonic smoother or the values $T_i$, $i=1,2$ of the two rectangular smoothers that compose the trapezoidal filter, the shape of the output trajectory is completely determined.
In particular, with the trapezoidal smoother it is possible to impose desired bounds on the velocity and acceleration of the output trajectory  by assuming
\be
T_1 = \frac{h}{\vmax}\hspace{15mm}
T_2 = \frac{\vmax}{\amax}
\label{eq:TrapezoidalT}
\ee
where $h$ denotes the amplitude of the step reference input (and consequently the amplitude of the desired displacement), $\vmax$ and $\amax$ are the limit values of velocity and acceleration, respectively. Note that, in this way, the minimum-time trajectory compliant with the given kinematic constraints is obtained.
 On the other side, it is convenient to use the harmonic smoother with the purpose of suppressing possible residual vibrations of the plant that must track the trajectory. It is possible to prove that if a system is characterized by a resonant frequency at $\omega_n$, residual vibrations are completely suppressed by setting the value of the time-constant $T$ as \be T = 3\frac{\pi}{\omega_n},\label{eq:HarmonicT}\ee
 see \cite{DampedSinusoidal}.
\begin{figure}[tb]
	\centering
	\psfrag{Pos}[c][c][0.8][0]{ $p(t)$}
	\psfrag{t}[c][c][0.8][0]{ $t$}
	\psfrag{Vel}[c][c][0.8][0]{ $\dot p(t)$}
	\psfrag{Acc}[c][c][0.8][0]{ $\ddot p(t)$}
	\psfrag{TTTTTTTT}[l][l][0.45][0]{ Trapezoidal}
	\psfrag{HHHHHHHH}[l][l][0.45][0]{ Harmonic}
	\includegraphics[width=0.7\columnwidth]{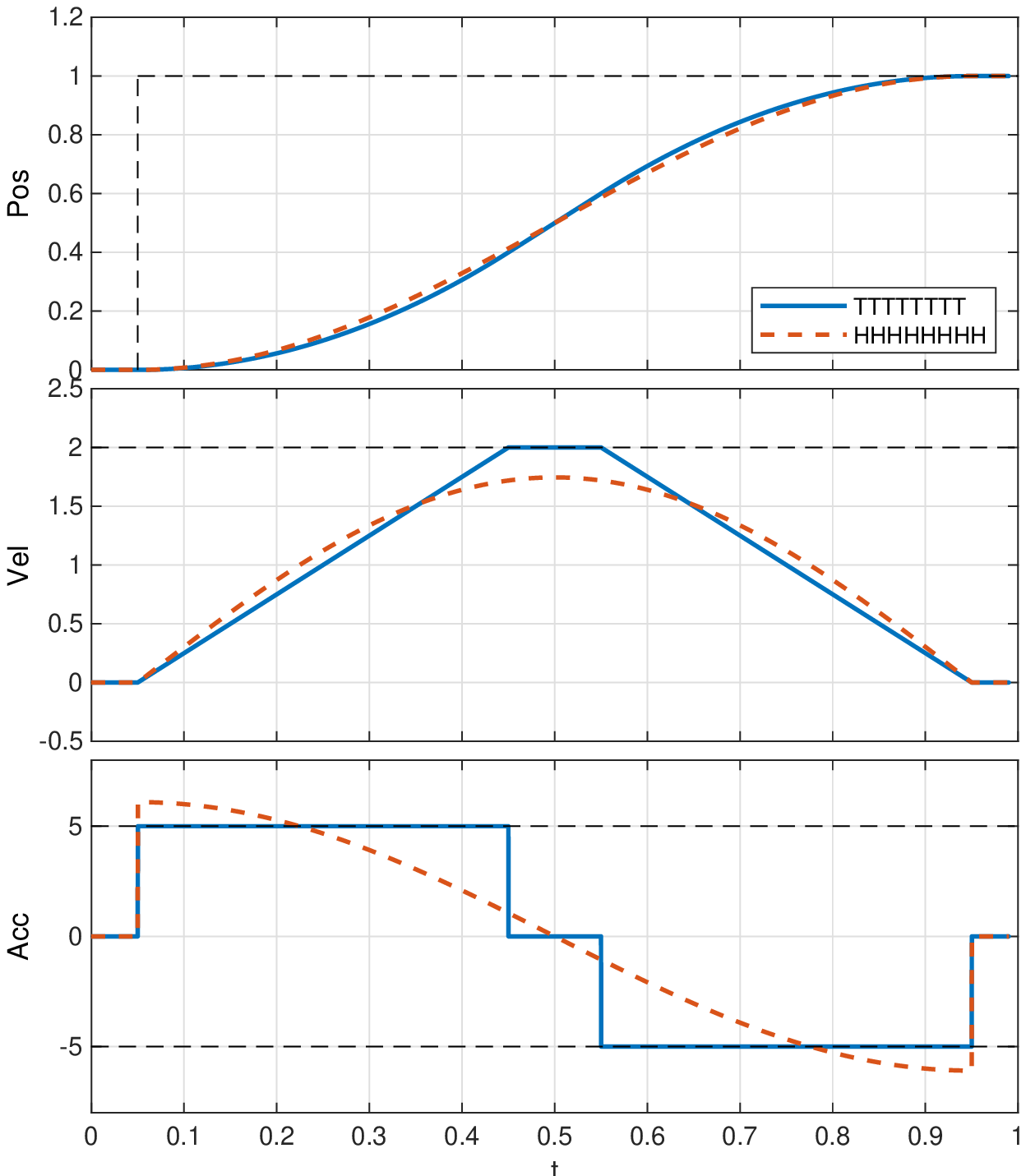}\\
{\small (a)}\\[4mm]
	\psfrag{W}[c][c][0.8][0]{ $\omega_n$}
	\psfrag{w}[c][c][0.6][0]{ $\omega_n$}
	\psfrag{A}[c][c][0.6][0]{ $1/\sqrt{2}$}
	\psfrag{B}[c][c][0.8][0]{ $\omega_T$}
	\psfrag{T}[l][l][0.45][0]{ Trapezoidal}
	\psfrag{H}[l][l][0.45][0]{ Harmonic}
	\psfrag{X}[c][c][0.8][0]{ $\omega$}
	\psfrag{x}[c][c][0.6][0]{ $\omega$}
	\psfrag{Y}[c][c][0.8][0]{ $|H(j\omega)|$}
	\psfrag{y}[c][c][0.6][0]{ $|H(j\omega)|$}
	\includegraphics[width=0.9\columnwidth]{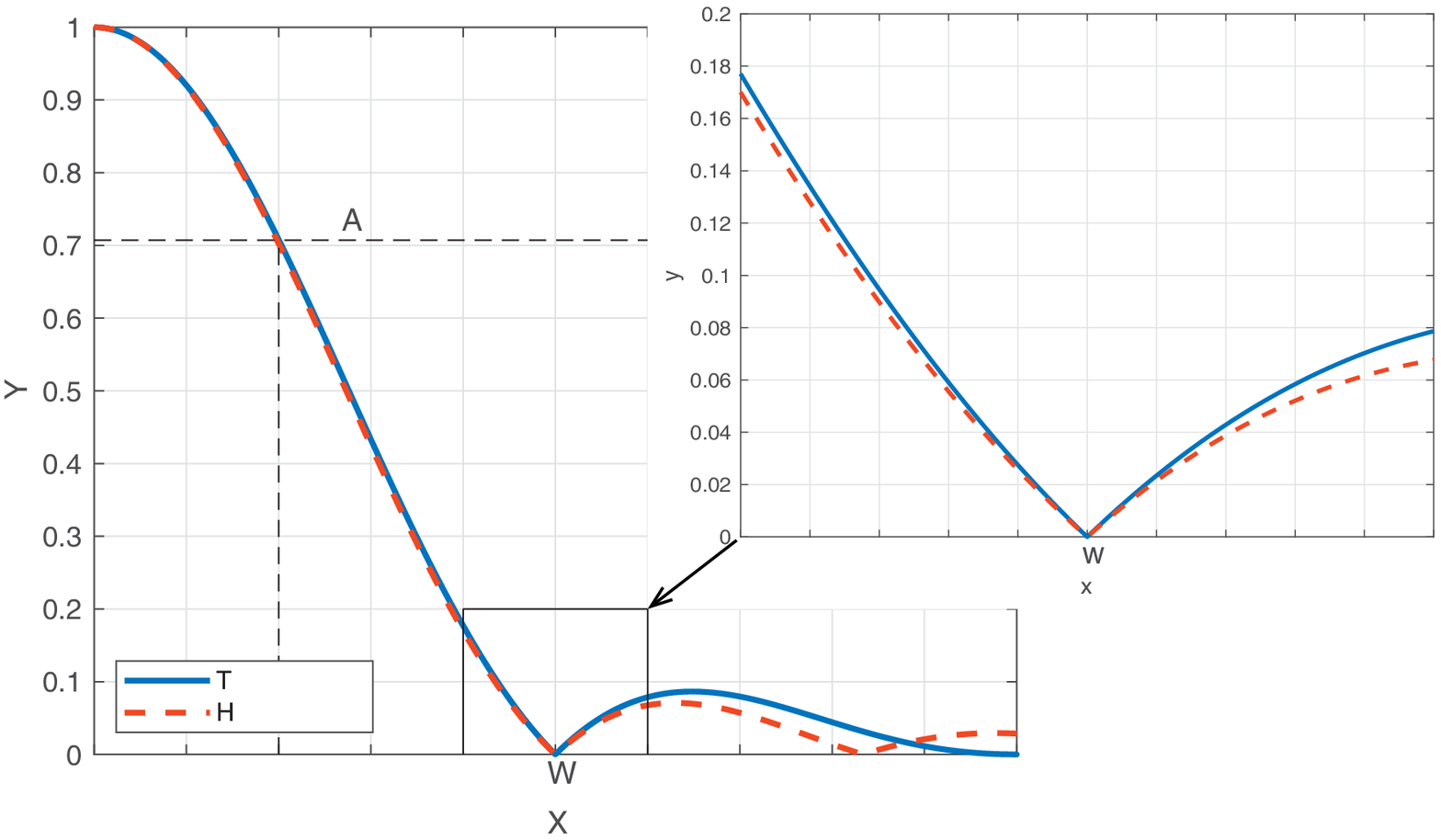}\\[-1mm]
{\small (b)}
	\caption{\label{fig:TrapVsHarm} Step response (a) and frequency response (b) of trapezoidal and harmonic smoothers.}
\end{figure}

In Fig. \ref{fig:TrapVsHarm}.a, we compare the step responses of a trapezoidal smoother and a harmonic smoother with the same duration. The time constants $T_i$ of the trapezoidal smoother are computed using (\ref{eq:TrapezoidalT}) with $h=1$, $\vmax=2$, and $\amax=5$, while the parameter $T$ of the harmonic smoother is computed as $T = T_1+T_2$. Both output signals have limited accelerations, but the harmonic smoother does not comply with the given constraints.
In Fig. \ref{fig:TrapVsHarm}.b, we consider the magnitude of the frequency response $|H(j\omega)|$ of the two smoothers. For the trapezoidal smoother, $T_1 = \frac{2\pi}{\omega_n}$ and $T_2 = \frac{\pi}{\omega_n}$, while for the harmonic smoother, $T$ is computed using (\ref{eq:HarmonicT}). In this way, the delay caused by the two types of smoothers is exactly the same, and both have a frequency response with zero magnitude at $\omega=\omega_n$.
Note that $|H(j\omega)|$ is proportional to the Percent Residual Vibration (PRV) induced by the filtered signal when it is applied to a resonant plant completely undamped, and whose resonant frequency differs from the nominal value $\omega_n$. Refer to \cite{DampedSinusoidal} for more details. Therefore, $|H(j\omega_n)| = 0$ implies that a residual vibration exactly at $\omega_n$ is completely suppressed, while for values $\omega$ different from  $\omega_n$, the smaller $|H(j\omega)|$ is, the higher its capability of reducing the amplitude of vibrations. Accordingly, Fig. \ref{fig:TrapVsHarm}.b shows the superiority of the harmonic smoother over the trapezoidal one in suppressing residual vibration, as the magnitude of its frequency response is smaller over the entire range of frequencies.
Interestingly enough, both smoothers have a frequency response with strong low-pass characteristics (with a cut-off frequency  $\omega_T\approx \frac{2}{5}\omega_0$ as shown in \Fig{TrapVsHarm}.b), which has two important consequences:
\begin{itemize}
\item The sloshing modes characterized by natural frequencies higher than $\omega_n$ are effectively reduced, even if the smoother was not specifically designed for them.
\item High-frequency noise superimposed on the input signal is significantly reduced. This property is particularly useful for applications where spurious signals affect the desired reference, and acceleration needs to be computed. It's worth noting that while the example shown in Fig. \ref{fig:TrapVsHarm}.a demonstrates the use of a step function as input for both smoothers, they can be applied to any reference signal produced, for example, by a human being according to a direct telemanipulation scheme.
\end{itemize}
As a final remark, it's worth noticing that a proper implementation of the smoother provides not only the filtered output but also its first and second derivative, without the use of differentiators. This is illustrated in detail in Sec. \ref{ssec:P2P_liquid}.

\section{Optimal feed-forward control for nonprehensile 3D manipulation} \label{sec:feedforward}  
\begin{figure*}[tbhp]
	\centering
	\psfrag{O}[tc][l][0.7][0]{\begin{tabular}{c@{}c@{}}Angles computation\\ according to (\ref{eq:theta}) and (\ref{eq:varphi})\end{tabular}}
	\psfrag{A}[c][c][0.7][0]{$ \beta \!=\!\!  \tan^{-1}\!\!\left(\!\!\frac{\sqrt{\ddot {\hat x}^2+\ddot {\hat y}^2}}{g + \ddot {\hat z}}\!\!\right)$}
	\psfrag{a}[c][c][0.7][0]{$ \varphi\!=\!\!   \pi + \mathrm{atan2}(\ddot {\hat y},\ddot {\hat x})$}
	\psfrag{R}[c][c][0.7][0]{${\bf R}\big(\beta(t), \varphi(t)\big)$}
	\psfrag{q}[c][c][0.7][0]{$\q(t)$}
	\psfrag{r}[c][c][0.8][90]{Robot controller}
	\psfrag{F}[c][c][0.9][0]{ $H(s)$}
	\psfrag{h}[c][c][0.7][0]{ \hspace{-2.5mm}${^0\T_{\mbox{\footnotesize CoR}}}(t)$}
	\psfrag{j}[c][c][0.7][0]{ \hspace{-2mm}${\cdot\,^{\mbox{\footnotesize F}}\T^{-1}_{\mbox{\footnotesize CoR}}}$}
	\psfrag{k}[c][c][0.7][0]{\hspace{-3mm} \mbox{\parbox[c]{2cm}{\centering Inverse\\ Kinematics}}}
	\psfrag{p}[c][c][0.6][0]{ ${\bf p}(t)\!=\!\!  \left[\!\!\!\begin{array}{c} x(t)\\y(t)\\z(t)\end{array}\!\!\!\right]$}
	\psfrag{P}[c][c][0.5][0]{ $\hat{\bf p}(t)\!=\!\!  \left[\!\!\!\begin{array}{c} \hat x(t)\\ \hat y(t)\\ \hat z(t)\end{array}\!\!\!\right]$}
	\psfrag{d}[c][c][0.5][0]{ $\ddot {\bf p}(t)\!=\!\!  \left[\!\!\!\begin{array}{c} \ddot x(t)\\ \ddot y(t)\\ \ddot z(t)\end{array}\!\!\!\right]$}
	\psfrag{D}[c][c][0.5][0]{ $\ddot {\hat{\bf p}}(t)\!=\!\!  \left[\!\!\!\begin{array}{c} \ddot {\hat x}(t)\\ \ddot {\hat y}(t)\\ \ddot {\hat z}(t)\end{array}\!\!\!\right]$}
\psfrag{v}[c][c][0.5][0]{ $\dot {\hat{\bf p}}(t)$}
	\includegraphics[width=1.9\columnwidth]{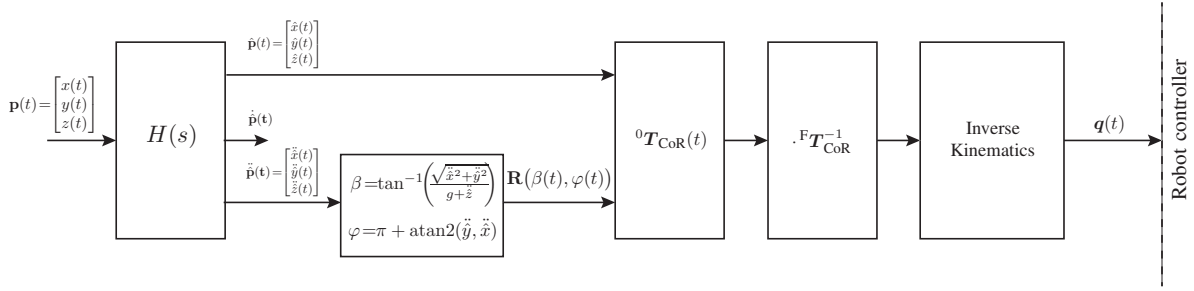}\\
	\caption{\label{fig:algorithm} Feed-forward controller for nonprehensile 3D manipulation.}
\end{figure*}

\begin{figure}[tb]
	\centering
	\psfrag{a}[c][c][1][0]{ $x$}
	\psfrag{b}[c][c][1][0]{ $y$}
	\psfrag{c}[c][c][1][0]{ $z$}
	\psfrag{d}[c][c][1][0]{ $\varphi$}
	\psfrag{e}[c][c][1][0]{ $\beta$}
	\psfrag{f}[c][c][1][0]{ $z_P$}
	\psfrag{l}[c][l][1][0]{ $l$}
	\psfrag{m}[c][c][1][0]{ $m$}
	\includegraphics[width=0.4\columnwidth]{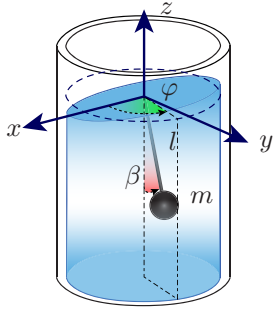}
	\caption{\label{fig:sph_pend_mod} Spherical pendulum modelling the sloshing dynamics in a liquid-filled vessel that moves along a 3D trajectory.}
\end{figure}

The general structure of the feed-forward controller, that assures a safe handling
of both solid and liquid materials without any fixturing mechanism,
is shown in \Fig{algorithm}.
A smoother is fed with the desired position, which can be a simple constant value denoting the goal in point-to-point motions or a complex reference trajectory provided e.g. by a human operator in a telerobotic architecture as described in \cite{Selvaggio2022}. The structure of the smoother and the characteristics of the reference signal depends on the considered application.
The output is a filtered trajectory $\hat p(t) = [\hat x(t),\,\hat y(t),\,\hat z(t)]^T$, with bounded (and known) acceleration that can be used to  is  used for constructing the orientation trajectory of the robot manipulator  with the purpose of aligning the container with the (equilibrium) angular position of the virtual pendulum that otherwise will be caused by the acceleration $\ddot{\hat p}(t)$, see remark \ref{rem:PendAlign}. As shown in \Fig{sph_pend_mod}, the spherical pendulum configuration, describing the sloshing phenomenon in the 3D space,  can be fully described by means of the  angles $(\beta,\,\varphi)$. The dependence of these angles from the linear acceleration imposed to the vessel can be analytically deduced \cite{Moriello2017}, i.e.
\begin{align}
\beta =&  -\tan^{-1}\left(\frac{\sqrt{\ddot {\hat x}^2+\ddot {\hat y}^2}}{g + \ddot {\hat z}}\right)  \label{eq:theta} \\[2mm]
\varphi=&  \,\,\,\pi + \mathrm{atan2}\,\,(\ddot {\hat y},\ddot {\hat x}) \label{eq:varphi}
\end{align}
where $\mathrm{atan2}$ is the four quadrant inverse tangent.
Accordingly the desired orientation for the object/vessel containing the liquid is
\be {\bf R}(\beta, \varphi) = \mathrm{Rot}_z (\varphi)\,\mathrm{Rot}_y (\beta)\,\mathrm{Rot}_z (-\varphi). \label{eq:Rotation} \ee
It is worth to noticing that the term $\mathrm{Rot}_z (-\varphi)$ not only compensates for the initial rotation of the transported object $\mathrm{Rot}_z (\varphi)$ but also offers an additional advantage: when $\ddot {\hat y} = \ddot {\hat x} = 0$,  the angle $\varphi$ in (\ref{eq:varphi}) is not well defined; however, since $\beta=0$, ${\bf R}(\theta, \varphi) = \mathrm{Rot}_z (\varphi)\,\mathrm{Rot}_z (-\varphi) = \I_3$, being $\I_3$ the $3-$by$-3$ identity matrix.\\ Obviously, when the motion is restricted to the $x-z$ plane (being $\ddot y=0$) as in the simplified example of \Sec{model}, the rotation imposed by (\ref{eq:Rotation}) with (\ref{eq:theta}) and (\ref{eq:varphi}) coincides with (\ref{eq:AngleOfRotation}).\\
Finally,  it is necessary to specify the point where the rotation must take place, that according to the observations  of \Sec{model}, and in particular remarks
\ref{rem:PendAlign} and \ref{rem:CoR}, changes on the basis of the material to be handled. In the case of a liquid in a container, the Center of Rotation (CoR) should be located in the center of the swinging mass $m$, while in the case of a solid object, the CoR should be placed in the Center of Mass (CoM) of the body.\\

In conclusion, the vector of filtered trajectory $\hat {\bf p}(t)$ and the rotation matrix ${\bf R}(\beta(t),  \varphi(t))$ are organized into the homogeneous transformation matrix
\be
^0\T_{\mbox{\footnotesize CoR}}(t) = \left[\begin{array}{c c}{\bf R}( \beta(t),  \varphi(t)) & \hat {\bf p}(t)\\ 0\hspace{0.6cm} 0\hspace{0.6cm} 0 &1 \end{array}\right]
\label{eq:T_0_CoR}
\ee
that provides the desired configuration of the reference frame attached to CoR with respect to the world reference frame $\mathcal{F}_0$. Then, it is necessary  to take into account the relative position of the CoR with respect to the robot flange by means of the constant (at least for a specific object) matrix $^{\mbox{\footnotesize F}}\T_{\mbox{\footnotesize CoR}}$: $$^{\mbox{\footnotesize 0}}\T_{\mbox{\footnotesize F}}(t) = \, ^{\mbox{\footnotesize 0}}\T_{\mbox{\footnotesize CoR}}(t)\cdot \, ^{\mbox{\footnotesize F}}\T_{\mbox{\footnotesize CoR}}^{-1}.$$
Finally, the instant configuration of the robot flange $^{\mbox{\footnotesize 0}}\T_{\mbox{\footnotesize F}}(t)$ is  processed using the inverse kinematics to derive the  joint trajectories $\q(t)$ that the robot manipulator must track.\\

The selection of the proper smoother $H(s)$ for a given application is the last  problem to be addressed. The goal is to minimize either the total duration of the trajectory in the case of point-to-point motions, or the additional delay imposed by the filter when a generic input trajectory is given. Four different scenarios are therefore possible, which are analyzed below.

\subsection{Point-to-point trajectory for solid object manipulation} \label{ssec:P2P_solid}
Since the lateral acceleration is the disturbance that affects the dynamics of the object resting on the tray, basic considerations suggest to minimize this quantity. Given a point-to-point motion, from the current location $\p_0 = [x_0,\,y_0,\,z_0]^T$ to the goal position $\p_1 = [x_1,\,y_1,\,z_1]^T$,  it is possible to prove that the trajectory that minimizes the acceleration given the duration $T$, or conversely the duration for a  given bound $\amax$, is the so-called {\it triangular velocity} trajectory, characterized by a bang-bang profile of the acceleration.
This can be obtained by filtering a step input (from $\p_0$ to $\p_1$) with a special type of trapezoidal smoother with
\[
T_1 = T_2\,\,\,\Rightarrow\,\,\, \vmax = \sqrt{h\,\amax}
\]
where $h=\|\p_1 -\p_0\|$.
The total duration of the output trajectory is
\be T = T_1+T_2 = 2\sqrt{\frac{h}{\amax}}\,\,\, \Rightarrow \,\,\,\amax = \frac{4\,h}{T^2}
\label{eq:MinAccSmoother}\ee
By exploiting the relationship between $T$ and $\amax$ ($=\|\ddot \p(t)\|$), it is possible to deduce the limit value of the trajectory's duration that assures safe transportation even in the absence of tilting compensation (see remark \ref{rem:FrictionLimit}), namely finding $T$ that
\be
\mbox{maximizes } \|\ddot \p(t)\|^2 = \ddot x^2 + \ddot y^2 + \ddot z^2  \label{eq:MaxProblem}\ee
\be
\mbox{subject to } \frac{\sqrt{\ddot x^2 + \ddot y^2}}{g+\ddot z}\le\mu \label{eq:Constraints}\ee
If the motion is decomposed into a vertical and a horizontal component, i.e. $\p_o$ and $p_v$ characterized by  \beann h_o &=& \sqrt{(x_1-x_0)^2+(y_1-y_0)^2}\\
h_v &=&|z_1-z_0|\eeann  respectively, the constraint (\ref{eq:Constraints}) for a triangular velocity  trajectory becomes, in the worst case\footnote{As a worst case, it is assumed that $\ddot z = -\amax = - \frac{4\,h_v}{T^2}$.},
\[
\frac{ \ds  \frac{4\,h_o}{T^2}}{\ds g- \frac{4\,h_v}{T^2}}\le \mu
\]
and accordingly, the duration of the motion is
\be
T\ge T^\star = 2\sqrt{\frac{h_o+\mu\, h _v}{\mu\,g}}.
\label{eq:MinTimeFlatMotion}
\ee
Note that $T^\star$ is a lower bound for the duration of a point-to-point motion that cannot be exceeded in any way if the object's stability on the tray is ensured solely by friction, without tilting compensation. As stated in Remark \ref{rem:CoR}, the orientation compensation (\ref{eq:Rotation}) removes this bound on $T$ because, in principle, any lateral acceleration can be compensated by this mechanism. Therefore, a trapezoidal smoother can be adopted in this case instead of a triangular one to take into account bounds on the maximum velocity and acceleration that the robot can achieve. Accordingly, the parameters of the smoother can be computed according to (\ref{eq:TrapezoidalT}). However, a single second-order smoother is not sufficient because the orientation compensation requires a trajectory with a degree of continuity higher than one to achieve limited angular velocities and accelerations. Specifically, due to the relationship between the trajectory's accelerations and the angles $\beta$, $\varphi$ in (\ref{eq:theta})-(\ref{eq:varphi}), it is necessary that $\ddot{\hat{x}}$, $\ddot{\hat{y}}$, and $\ddot{\hat{z}}$ are of class $\mathcal{C}^1$. This can be achieved by combining the smoother, which determines the basic point-to-point motion, with another second-order smoother. This additional filter could be, for example, a triangular smoother with parameters $T_1=T_2$, such that the angular velocity and acceleration are below the desired values. Due to the nonlinear relationships (\ref{eq:theta})-(\ref{eq:varphi}), these values cannot be expressed analytically in terms of $T_1$ and $T_2$, but from (\ref{eq:MinAccSmoother}), it follows that the larger the $T_i$, $i=1,2$, the lower the maximum (angular) acceleration. Therefore, the selection of the proper values of parameters $T_i$ should be carried out in the field by imposing angular speeds and accelerations that the robot is able to reach. No other consideration is instead linked to the stability of the object on the tray, which is guaranteed in any case, see Remark \ref{rem:CoR}.

\subsection{Point-to-point trajectory for liquids transportation}
\label{ssec:P2P_liquid}
\begin{figure}[tb]
	\centering
 \psfrag{w}[l][l][0.8][0]{Im\{s\}}
 \psfrag{s}[c][c][0.8][0]{Re\{s\}}
 \psfrag{A}[l][l][0.8][0]{$+j\frac{3}{2}\frac{2\pi}{T}$}
 \psfrag{a}[l][l][0.8][0]{$-j\frac{3}{2}\frac{2\pi}{T}$}
 \psfrag{Z}[l][l][0.8][0]{$+j\frac{1}{2}\frac{2\pi}{T}$ }
 \psfrag{z}[l][l][0.8][0]{$-j\frac{1}{2}\frac{2\pi}{T}$ }
 \psfrag{S}[l][l][0.8][0]{$\sigma$}
 \psfrag{T}[l][l][0.8][0]{$+j\frac{(2k+1)}{2} \frac{2\pi}{T}$ }
 \psfrag{t}[l][l][0.8][0]{$-j\frac{(2k+1)}{2} \frac{2\pi}{T}$ }
\includegraphics[width=0.6\columnwidth]{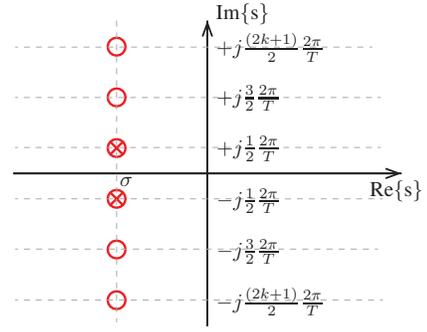}\\
	\caption{\label{fig:Pzmap_H_s} Pole-zero map of the damped harmonic smoother $H(s)$: $\sigma$ and $T$ are the free parameters that appear in (\ref{eq:DampedSinusoidalFilterCTF}).}
\end{figure}

In this case, the goal of minimizing the maximum acceleration is subordinated to the need to cancel the residual vibrations on the pendulum that models the sloshing phenomenon. As a matter of fact, as shown in Sec. \ref{ssec:SloshingModel}, see equations (\ref{eq:PendDynamicsLinear}) and (\ref{eq:SloshingLinMod}), the liquid in the container behaves like a second-order system with given natural frequency $\omega_n$ and damping ratio $\delta$. Accordingly, for a point-to-point motion, the harmonic smoother is preferred to the trapezoidal one, as it guarantees greater robustness with respect to the problem of residual vibration suppression, having a lower PRV about the nominal value of $\omega_n$.\\
In particular, the so-called {\it damped harmonic smoother} is adopted \cite{DampedSinusoidal}, i.e. a smoother obtained by modifying the basic harmonic smoother in (\ref{eq:HarmonicSmoother}) to take into account the damping of the vibrating system and  whose analytical expression is
\begin{equation}
  H(s)  = \frac{\sigma^2 +  \left(\frac{\pi}{T}\right)^2}{1+e^{\sigma\, T}}  \frac{1+e^{-sT}e^{\sigma\, T}}{(s - \sigma)^2 + \left(\ds \frac{\pi}{T} \right)^2}.
  \label{eq:DampedSinusoidalFilterCTF}
\end{equation}
where $\sigma$ and $T$ are freely selectable constant parameters. 
The smoother is characterized by the pole-zero map of \Fig{Pzmap_H_s}. The cancellation of the oscillating dynamics  described by (\ref{eq:SloshingLinMod})  can be obtained by assuming
\be
\sigma = -\delta\omega_n, \hspace{5mm} T = \frac{3}{2}\frac{2\pi}{\omega_n\sqrt{1-\delta^2}}.
\label{eq:HarmonicSmootherParameters}
\ee
Since the parameters of the filter are deduced from the values $\delta$ and $\omega_n$ that characterize the system, having a reliable model becomes extremely important for effectively suppressing sloshing.\\
\begin{figure}[tb]
	\centering
	\psfrag{a}[c][c][0.7][0]{$\ds K$}
	\psfrag{c}[c][c][0.7][0]{$\ds e^{-sT}$}
	\psfrag{e}[c][c][0.6][0]{$\ds \sigma^2\!\!+\!\!\left(\!\frac{\pi}{T}\!\right)^2$}
	\psfrag{f}[c][c][0.6][0]{$\ds -2\sigma$}
	\psfrag{b}[c][c][0.7][0]{$\ds e^{\sigma\, T}$}
	\psfrag{s}[c][c][0.7][0]{$\ds \frac{1}{s}$}
	\psfrag{p}[c][c][0.7][0]{$\ds p(t)$}
    \psfrag{q}[c][c][0.7][0]{$\ds \hat p(t)$}	
    \psfrag{d}[c][c][0.7][0]{$\ds \dot{\hat p}(t)$}	
    \psfrag{D}[c][c][0.7][0]{$\ds \ddot{\hat p}(t)$}
    \psfrag{1}[c][c][0.7][0]{Two-impulse IS}
     \psfrag{2}[c][c][0.7][0]{Second order filter}
\includegraphics[width=1\columnwidth]{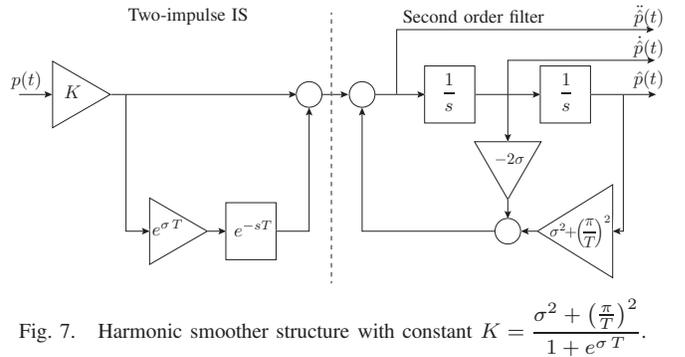}\\
	\caption{\label{fig:H_s} Harmonic smoother structure with constant $\ds K=\frac{ \sigma^2 + \left(\frac{\pi}{T}\right)^2}{1+e^{\sigma\, T}} $.}
\end{figure}
The structure of the filter expressed in the controllable canonical form, reported in \Fig{H_s}, allows to obtain not only the filtered output but also its first and second derivatives, without the need for an explicit differentiation, as mentioned in Sec. \ref{sec:Smoothers}. \\
As in the case of point-to-point motions for solid objects, the tilting compensation of the lateral acceleration requires the continuity of its first derivative. This can be achieved by combining the damped harmonic smoother with an additional triangular smoother, whose parameters $T_i$ must be large  enough to produce feasible  angular velocities and accelerations.
\subsection{Complex trajectories for solid object manipulation} \label{ssec:complex_solid}
When the input trajectory is not a simple step signal specifying the final position, as in point-to-point motions, but a complex motion, possibly unknown in advance, a single second-order smoother is sufficient for the proposed application. It can be assumed that the input signal already has limited acceleration, such as trajectories defined by parametric curves (e.g., cubic splines that are of class $\mathcal{C}^2$) or motions commanded by a human operator via direct teleoperation, which are characterized by continuous acceleration. In this case, the smoother has two objectives: to impose a bound on the higher order derivatives of the acceleration to achieve tilting compensation with limited angular velocities and accelerations, and simultaneously estimate the value of linear acceleration without explicit differentiation for analytical calculation of angular compensation. This is particularly useful when the input signal is provided by a human operator and is therefore affected by some level of noise, caused by both the sensors used for position detection and natural tremors that affect human beings. As mentioned in Section \ref{sec:Smoothers}, the smoothers exhibit low-pass characteristics, allowing their parameters to be selected to effectively reject noise. For example, in the case of a triangular smoother\footnote{The bandwidth of the rectangular smoother with a generic time constant $T_i$, that compose a triangular smoother, is approximately $2/T_i$.}, the larger the values of the two parameters, $T_1$ and $T_2$, the narrower the filter's bandwidth becomes, resulting in more effective noise filtering. However, it is important to consider that the smoother introduces a delay of exactly $T_1 + T_2$ between the input signal and the filtered output. Consequently, it is advantageous to keep these values as small as possible and find a trade-off between the different requirements based on the specific application.
\subsection{Complex trajectories for liquids manipulation} \label{ssec:complex_liquid}
Applications that require robots to handle liquid materials along complex trajectories are subject to the same constraints as those for solid materials. Therefore, the considerations outlined above remain valid. However, when dealing with liquids, there is an additional need to suppress sloshing. In this case, the harmonic smoother is the best solution due to its superior robustness against residual vibrations. Unlike the triangular smoother suggested for solid objects, the parameters that characterize the harmonic smoother cannot be freely selected within a given range (determined by the bounds on acceleration, noise reduction, etc.). Instead, these parameters must be chosen based on the features of the liquid, using (\ref{eq:HarmonicSmootherParameters}).\\

The procedure for the selection of the smoothing filter $H(s)$ and the matrix $^{\mbox{\scriptsize F}}\T_{\mbox{\scriptsize CoR}}$ that define the proper control scheme of Figure \ref{fig:algorithm} in different application scenarios is outlined in Table \ref{tab:Summary}.
Note that the rotation matrix $^{\mbox{\scriptsize F}}{\bf R}{\mbox{\scriptsize x}}$ remains the same for any object/container and is based on an arbitrary assumption. Knowledge of the vector describing the location of the center of mass of a solid object, $^{\mbox{\scriptsize F}}\p{\mbox{\scriptsize CoM}}$, or the location of the equivalent mass $m$ of the liquid in the container, $^{\mbox{\scriptsize F}}\p_{\mbox{\scriptsize m}}$, requires information about the shape and mass distribution of the object, as well as additional sensors. Specifically, with a force sensor installed in the robot's flange, it is possible to estimate the centroid's location on the tray of the object/container. Therefore, only the knowledge of the object's center of mass height or the liquid level\footnote{It is worth noting that the vertical position of the mass $m$ can be easily deduced from the liquid surface position by subtracting the length $l$ of the equivalent pendulum, computed from $\frac{g}{l} = \omega_n^2$ (see equations \ref{eq:PendDynamicsLinear} and \ref{eq:SloshingLinMod}).}  is necessary.

\begin{table*}
\begin{center}
\renewcommand{\arraystretch}{1.5}
\begin{tabular}{c|p{8cm}|p{8cm}|}
  \multicolumn{3}{r}{   \textbf{Point-to-Point trajectories} \hspace{50mm} \textbf{Complex trajectories} \hspace{20mm}~}\\
   \cline{2-3}
   \rotatebox[origin=c]{90}{\textbf{Solid objects~~~~~~}} & \parbox{8cm}{\vspace{2mm}{\bf Inputs:} goal position $^0\p_1$, center of mass of the object $^t\p_{\tiny CoM}$, maximum cartesian velocity $\vmax$ and acceleration $\amax$ of the robot\\[2mm]
   {\bf Design of Filter $H(s)$:} trapezoidal smoother with parameters computed according to (\ref{eq:TrapezoidalT}) + triangular smoother with free parameters\\[2mm]
   {\bf Definition of the transformation $^{\mbox{\scriptsize F}}\T_{\mbox{\scriptsize CoR}}$:} 
 \be
^{\mbox{\scriptsize F}}\T_{\mbox{\scriptsize CoR}} = \left[\begin{array}{c c}^{\mbox{\scriptsize F}} {\bf R}_{\mbox{\scriptsize Obj}}  &  ^{\mbox{\scriptsize F}}\p_{\mbox{\scriptsize CoM}}\\ {\bf 0} &1 \end{array}\right]
\nonumber 
\ee
where $^{\mbox{\scriptsize F}}{\bf R}_{\mbox{\scriptsize Obj}}$ is the constant rotation matrix describing the orientation of the object with respect to the flange reference frame and $^{\mbox{\scriptsize F}}\p_{\mbox{\scriptsize CoM}}$ denotes the position of the center of mass of the object in the same reference frame\\[2mm]
   }& \parbox{8cm}{\vspace{2mm}{\bf Inputs:} desired reference trajectory $^0\p(t)$, center of mass of the object $^t\p_{\tiny CoM}$\\[2mm]
   {\bf Design of Filter $H(s)$:}  triangular smoother with free parameters\\[2mm]
   {\bf Definition of the transformation $^{\mbox{\scriptsize F}}\T_{\mbox{\scriptsize CoR}}$:} 
 \be
^{\mbox{\scriptsize F}}\T_{\mbox{\scriptsize CoR}} = \left[\begin{array}{c c}^{\mbox{\scriptsize F}} {\bf R}_{\mbox{\scriptsize Obj}}  &  ^{\mbox{\scriptsize F}}\p_{\mbox{\scriptsize CoM}}\\ {\bf 0} &1 \end{array}\right]
\nonumber 
\ee
where $^{\mbox{\scriptsize F}}{\bf R}_{\mbox{\scriptsize Obj}}$ is the constant rotation matrix describing the orientation of the object with respect to the flange reference frame and $^{\mbox{\scriptsize F}}\p_{\mbox{\scriptsize CoM}}$ denotes the position of the center of mass of the object in the same reference frame.\\[2mm]
   } \\
   \cline{2-3}
   \rotatebox[origin=c]{90}{\textbf{Liquid materials}} & 
\parbox{8cm}{\vspace{2mm}{\bf Inputs:} goal position $^0\p_1$, natural frequency $\omega_n$ and damping ratio $\delta$ of the first sloshing mode, location of   the equivalent pendulum bob $^t\p_{\tiny m}$, maximum cartesian velocity $\vmax$ and acceleration $\amax$ of the robot\\[2mm]
   {\bf Design of Filter $H(s)$:}  trapezoidal smoother with parameters computed according to (\ref{eq:TrapezoidalT}) + damped harmonic smoother with parameters defined by (\ref{eq:HarmonicSmootherParameters})\\[2mm]
   {\bf Definition of the transformation $^{\mbox{\scriptsize f}}\T_{\mbox{\scriptsize CoR}}$:} 
 \be
^{\mbox{\scriptsize F}}\T_{\mbox{\scriptsize CoR}} = \left[\begin{array}{c c}^{\mbox{\scriptsize F}} {\bf R}_{\mbox{\scriptsize c}}  &  ^{\mbox{\scriptsize F}}\p_{\mbox{\scriptsize m}}\\ {\bf 0} &1 \end{array}\right]
\nonumber 
\ee
where $^{\mbox{\scriptsize F}}{\bf R}_{\mbox{\scriptsize c}}$ is the constant rotation matrix describing the orientation of the container with respect to the flange reference frame and $ ^{\mbox{\scriptsize F}}\p_{\mbox{\scriptsize m}}$ denotes the position of the  pendulum mass $m$ in the same reference frame.\\[2mm]
   }   
   & 
   \parbox{8cm}{\vspace{2mm}{\bf Inputs:} desired reference trajectory $^0\p(t)$, natural frequency $\omega_n$ and damping ration $\delta$ of the first sloshing mode, location of   the equivalent pendulum bob $^t\p_{\tiny m}$\\[2mm]
   {\bf Design of Filter $H(s)$:}  damped harmonic smoother with parameters defined by (\ref{eq:HarmonicSmootherParameters})\\[2mm]
   {\bf Definition of the transformation $^{\mbox{\scriptsize f}}\T_{\mbox{\scriptsize CoR}}$:} 
 \be
^{\mbox{\scriptsize F}}\T_{\mbox{\scriptsize CoR}} = \left[\begin{array}{c c}^{\mbox{\scriptsize F}} {\bf R}_{\mbox{\scriptsize c}}  &  ^{\mbox{\scriptsize F}}\p_{\mbox{\scriptsize m}}\\ {\bf 0} &1 \end{array}\right]
\nonumber 
\ee
where $^{\mbox{\scriptsize F}}{\bf R}_{\mbox{\scriptsize c}}$ is the constant rotation matrix describing the orientation of the container with respect to the flange reference frame and $ ^{\mbox{\scriptsize F}}\p_{\mbox{\scriptsize m}}$ denotes the position of the  pendulum mass $m$ in the same reference frame.\\[2mm]
   }
   \\
   \cline{2-3}
\end{tabular}
\end{center}
\caption{\label{tab:Summary} Computation of the parameters of the feed-forward control scheme of \Fig{algorithm} in different scenarios.}
\end{table*}


\section{Experimental validation}\label{sec:experiments}
Because of some issues in our lab, we are still in the process of concluding all the planned experiments. In the meantime, please refer to the videos available at the following URL: \url{https://sites.google.com/view/robotwaiter/}, where the proposed approach has been demonstrated.

\bibliographystyle{IEEEtran}
\bibliography{SloshingBib,NonPrehensileGrasp,RobotWaiter}
\end{document}